\documentclass[preprint,12pt]{elsarticle}

\usepackage{amssymb}
\usepackage{amsthm}
\usepackage{lineno}
\usepackage{bm}
\usepackage[colorlinks=true, pdfauthor=author]{hyperref}
\usepackage{wrapfig}
\usepackage{booktabs} 
\usepackage{algorithm}
\usepackage{algorithmic}
\usepackage{graphicx}
\usepackage{amsmath}
\usepackage{amssymb}
\usepackage{amsfonts}
\usepackage{multirow}
\usepackage{graphicx}
\usepackage{mathrsfs}
\usepackage{colortbl}
\usepackage{subfigure}
\usepackage{indentfirst}
\usepackage{ulem}
\usepackage{url}
\usepackage{verbatim}
\usepackage{rotating}
\usepackage{adjustbox}
\usepackage{wrapfig}
\usepackage{setspace}

\graphicspath{{./images/}}

\journal{arXiv}
\begin{document}

\begin{frontmatter}



\title{Parameter-Efficient Adapter Tuning for Tabular-Image Multimodal Learning}

\author[first]{Jiaqi Luo\corref{cor1}}
\cortext[cor1]{Corresponding author}
\ead{jqluo@suda.edu.cn}
\affiliation[first]{organization={School of Mathematical Sciences, Soochow University},
            addressline={No.1 Shizi Street}, 
            city={Suzhou},
            postcode={215006}, 
            state={Jiangsu Province},
            country={China}}

\begin{abstract}
Tabular-image multimodal learning aims to improve predictive modeling by jointly using structured tabular attributes and visual data. 
Although pretrained encoders provide strong modality-specific representations, full fine-tuning can be computationally expensive, while keeping encoders frozen may limit task-specific adaptation.
We propose the Tabular-Image Adapter (\textbf{TI-Adapter}), a modality-specific adapter-based fine-tuning framework for efficient multimodal adaptation. 
TI-Adapter freezes the pretrained tabular encoder and learns an adapter after the extracted tabular embedding, while adapting the image branch with embedding-level and bottleneck-level adapters instead of full fine-tuning. 
Experiments on 20 tabular-image datasets show that TI-Adapter achieves competitive or better predictive performance than full fine-tuning while using substantially fewer trainable parameters. 
Ablation studies further demonstrate the importance of adapter placement for balancing performance and practical efficiency.
\end{abstract}




\end{frontmatter}


\section{Introduction}

\label{s:intro}

Tabular-image multimodal learning is an emerging setting for predictive modeling in real-world applications~\cite{acosta2022multimodal,huang2020fusion,borsos2024predicting}. 
Image data capture visual patterns, while tabular data provide structured attributes. 
These two modalities often contain complementary information, and effectively integrating them can improve prediction over unimodal models.

A common strategy for tabular-image multimodal learning is to use pretrained encoders to extract modality-specific representations, followed by a fusion module and a prediction head. 
In the image branch, pretrained convolutional networks such as ResNet~\cite{he2016deep} are widely used due to their strong transferability. 
In the tabular branch, recent pretrained tabular models, such as TabPFN~\cite{hollmann2022tabpfn,hollmann2025accurate,grinsztajn2026tabpfn}, provide strong tabular representations through in-context learning. 

While such pretrained encoders offer useful starting points, adapting them efficiently to downstream multimodal tasks remains insufficiently studied.
The main challenge is that tabular and image encoders have different adaptation properties. 
Visual encoders can be fully fine-tuned, but updating the entire image backbone is computationally expensive and requires storing gradients and optimizer states for many parameters. 
Simply freezing the image encoder reduces cost, but may limit task-specific visual adaptation. 
On the tabular side, pretrained tabular models based on in-context learning are not naturally optimized through standard end-to-end fine-tuning, making direct parameter updates less straightforward. 
Therefore, tabular-image multimodal learning requires an adaptation strategy that respects the different characteristics of the two modalities.

Parameter-efficient fine-tuning (PEFT)~\cite{xu2026parameter,xin2024parameter} provides a natural direction for addressing this issue. 
Instead of updating all pretrained parameters, PEFT methods freeze most of the backbone and introduce a small number of trainable parameters for task-specific adaptation. 
Adapter-based methods are particularly suitable in this setting because they can be applied to final embeddings or intermediate feature maps without modifying the original encoder architecture~\cite{gao2024clip,chen2024conv}. 

However, most existing PEFT studies focus on language and vision models, while its role in tabular-image multimodal learning remains less explored.
Motivated by this gap, we propose the Tabular-Image Adapter (\textbf{TI-Adapter}), a modality-specific adapter-based fine-tuning framework for tabular-image multimodal learning. 
For the tabular branch, we keep the pretrained TabPFN encoder frozen and introduce an adapter after the extracted tabular embedding. 
For the image branch, we investigate two types of adapters: an embedding-level adapter placed after the final ResNet representation, and bottleneck-level convolutional adapters inserted after bottleneck blocks in selected ResNet layers. 
This design enables task-specific adaptation for both modalities while avoiding full encoder fine-tuning.
We evaluate TI-Adapter on 20 tabular-image datasets covering both classification and regression tasks. 
The results show that adapter-based fine-tuning achieves competitive or better predictive performance compared with full ResNet fine-tuning, while substantially reducing the number of trainable parameters. 
Our ablation study further shows that deeper adapter insertion can improve raw performance, but may increase GPU memory. 
Overall, TI-Adapter provides a practical and efficient alternative to full fine-tuning for adapting pretrained tabular and image encoders in multimodal learning.

\section{Related Work}
\label{s:rel}

\subsection{Tabular-image Multimodal Learning}

Tabular-image multimodal learning aims to integrate structured tabular attributes and visual information for improved prediction. 
Existing methods can be broadly grouped into supervised learning and self-supervised learning.

In supervised multimodal learning, a common strategy is to use separate encoders to extract modality-specific representations and then fuse the resulting embeddings for prediction. The image encoder is often initialized from a pretrained visual backbone, while the tabular encoder is typically a task-specific model trained from scratch. Several studies have explored this paradigm in medical applications. CLIP-Lung~\cite{lei2023clip} and Nodule-CLIP~\cite{sun2024nodule} incorporate language models to inject clinical knowledge into image representations. Spasov et al.~\cite{spasov2019parameter} combine MRI images and clinical variables for Alzheimer's disease prediction. Liu et al.~\cite{liu2023functional} integrate Support Vector Regression with a 3D ResNet for acute ischemic stroke prediction. Multi-TransSP~\cite{zheng2022multi} fuses CNN-based image features and Transformer-based tabular features for survival prediction. Xue et al.~\cite{xue2024ai} convert heterogeneous modalities into fixed-length embeddings and use a Transformer for dementia etiology prediction. 
More recently, Luo et al.~\cite{luo2025time} use TabPFN as a tabular encoder to produce robust tabular embeddings and compare it with several tabular encoders, including XGBoost~\cite{chen2016xgboost}, MLP, NCART~\cite{luo2024ncart}, and Transformer-based models~\cite{gorishniy2021revisiting}. MultiModalPFN~\cite{kim2026multimodalpfn} extends TabPFN to multimodal learning by encoding non-tabular inputs into embeddings, transforming them into tabular-like representations, and combining them with tabular features for TabPFN prediction.

Self-supervised tabular-image learning often relies on contrastive objectives, which pull matched tabular-image pairs closer in the embedding space and push unmatched pairs apart. The learned encoders can then be fine-tuned or directly used for downstream supervised tasks. MMCL~\cite{hager2023best} introduces contrastive learning to the tabular-image setting and demonstrates gains over unimodal and supervised baselines. Huang~\cite{huang2023multimodal} combines a tabular attention module with contrastive learning for Alzheimer's disease prediction. TIP~\cite{du2024tip} addresses missing tabular values and modality disparity by incorporating masked tabular reconstruction and tabular-image matching. SiTL~\cite{du2025stil} further investigates semi-supervised learning for tabular-image multimodal representation learning.


\subsection{Parameter-Efficient Fine-Tuning}

Parameter-efficient fine-tuning (PEFT) \cite{xu2026parameter, xin2024parameter} aims to adapt pretrained models to downstream tasks by updating only a small number of task-specific parameters. Instead of fine-tuning the entire backbone, PEFT methods usually freeze most pretrained weights and introduce lightweight trainable components, such as adapters and low-rank updates. This strategy has become increasingly important for large pretrained models, where full fine-tuning can be expensive in terms of memory, computation, and task-specific storage.

Adapter tuning is a representative PEFT strategy. Houlsby et al.~\cite{houlsby2019parameter} introduced bottleneck adapters, where small trainable modules are inserted into pretrained networks while the backbone remains fixed. Related methods, such as prefix tuning~\cite{li2021prefix} and prompt tuning~\cite{lester2021power}, adapt frozen language models by optimizing task-specific continuous vectors. Beyond language modeling, adapter-based methods have also been used in vision-language learning. CLIP-Adapter~\cite{gao2024clip} adapts frozen CLIP representations by inserting lightweight feature adapters into the visual and language branch and blending adapted features with the original pretrained features. Tip-Adapter~\cite{zhang2021tip} further reduces the training cost by constructing a training-free adapter for few-shot CLIP adaptation.

Another important PEFT method is low-rank adaptation. LoRA~\cite{hu2022lora} freezes pretrained weights and injects trainable low-rank matrices into selected layers, representing weight updates through low-rank decompositions. This greatly reduces the number of trainable parameters while maintaining competitive performance. Several extensions further improve LoRA's flexibility and efficiency. AdaLoRA~\cite{zhang2023adalora} dynamically allocates the parameter budget across weight matrices, whereas QLoRA~\cite{dettmers2023qlora} combines LoRA with three innovations to enable memory-efficient fine-tuning of large language models.

\section{Methodology}
\label{s:method}

\subsection{Overview}

\begin{figure}[!ht]
    \centering
    \includegraphics[width=\linewidth]{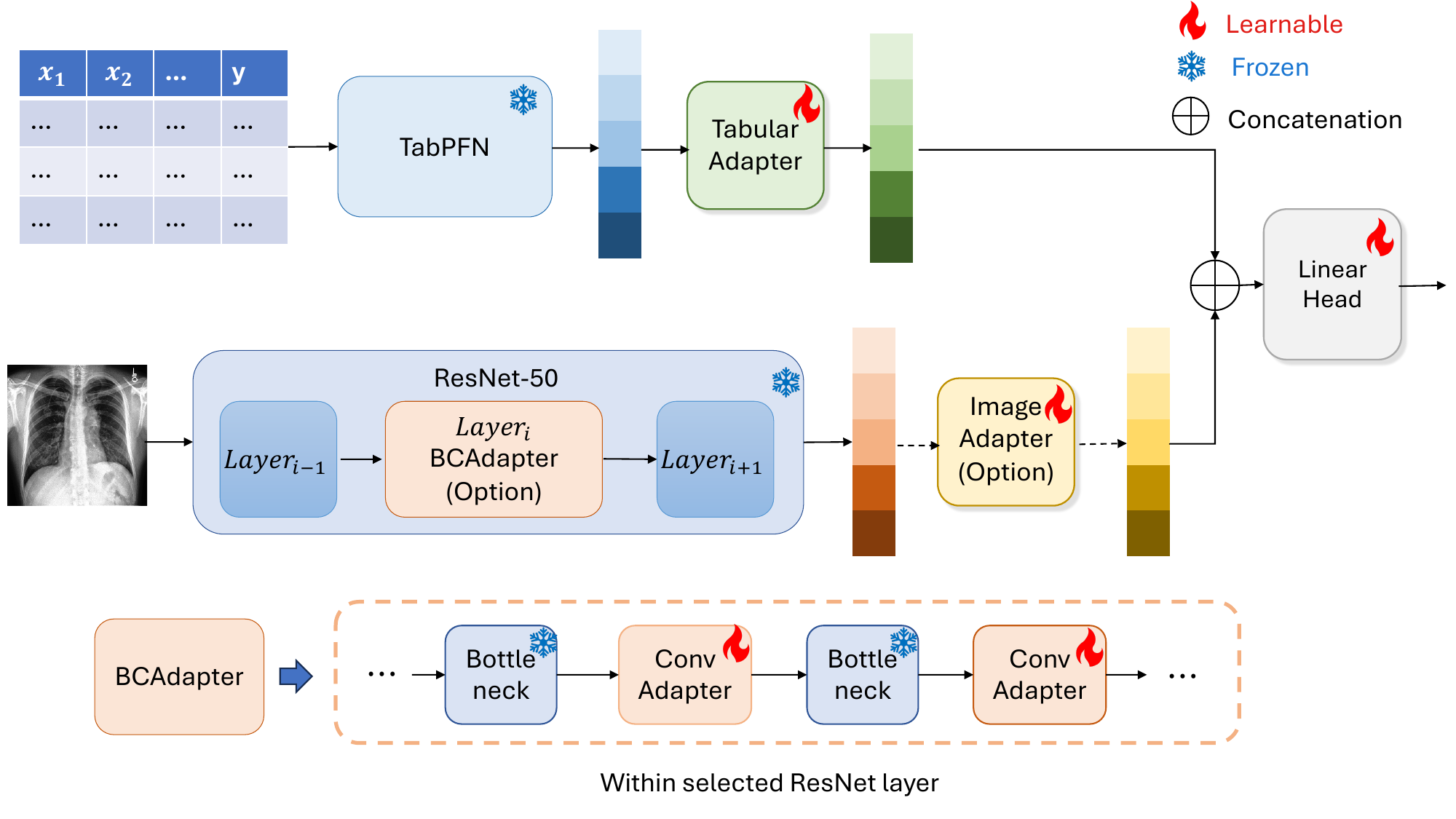}
    \caption{
    An overview of TI-Adapter. 
    Blocks marked with red flames are trainable, while blocks marked with blue snowflakes remain frozen during training. 
    The image branch supports two adaptation options: applying an embedding-level image adapter after the final ResNet representation, or inserting BCAdapters into selected ResNet layers.
    }
    \label{f.framework}
\end{figure}

Our proposed model, shown in Fig.~\ref{f.framework}, is designed for parameter-efficient tabular-image multimodal learning. The model consists of two pretrained modality-specific encoders and three trainable components: (1) a tabular adapter, (2) an image-side adapter, and (3) a linear predictor for classification or regression. The tabular encoder is implemented using TabPFN-v3 \cite{grinsztajn2026tabpfn} and the image encoder is implemented using ResNet-50 \cite{he2016deep}. The parameters in both encoders are kept frozen during the training.

\subsection{Tabular Adapter}
\label{s.tabular_adapter}

Given a tabular input $\mathbf{X} \in \mathbb{R}^{d}$, we use TabPFN-v3~\cite{grinsztajn2026tabpfn} as the tabular feature extractor. 
TabPFN is a transformer-based model trained offline to approximate Bayesian inference over synthetic tabular tasks. 
For a new dataset, it performs prediction through in-context learning by conditioning on a training set $\mathcal{D}_{\text{train}}$ and the test input, without updating its parameters:
\begin{equation}
    p(y_{\text{test}} \mid \mathbf{X}_{\text{test}}, \mathcal{D}_{\text{train}}).
\end{equation}
Here, $\mathcal{D}_{\text{train}}=\{(\mathbf{X}_i,y_i)\}_{i=1}^{N}$ denotes the training set, where $\mathbf{X}_i \in \mathbb{R}^{d}$ is a tabular feature vector and $y_i$ is the corresponding classification or regression label.

Since TabPFN relies on in-context learning, directly fine-tuning the entire TabPFN model is less straightforward than fine-tuning a standard neural backbone such as ResNet. 
Therefore, we keep the TabPFN encoder frozen and perform adaptation through an embedding-level adapter.

Given $\mathcal{D}_{\text{train}}$, the tabular embedding is extracted by the frozen TabPFN encoder:
\begin{equation}
\label{e.tabemb}
    \mathbf{E}_T = f_{\text{TabPFN}}(\mathbf{X}, \mathcal{D}_{\text{train}}),
\end{equation}
where $\mathbf{E}_T \in \mathbb{R}^{d_T}$ denotes the tabular representation and $d_T$ is the embedding dimension. 
We then introduce an adapter after the TabPFN embedding:
\begin{equation}
\label{e.tab_adapter}
    A_T(\mathbf{E}_T)
    =
    W_{T}^{\mathrm{up}}
    \sigma\left(
    W_{T}^{\mathrm{down}}\mathbf{E}_T+\mathbf{b}_{T}^{\mathrm{down}}
    \right)
    +\mathbf{b}_{T}^{\mathrm{up}},
\end{equation}
where $W_{T}^{\mathrm{down}} \in \mathbb{R}^{r_T \times d_T}$ projects the tabular embedding into a low-dimensional space, and $W_{T}^{\mathrm{up}} \in \mathbb{R}^{d_T \times r_T}$ projects it back to the original dimension. 
Here, $\mathbf{b}_{T}^{\mathrm{down}} \in \mathbb{R}^{r_T}$ and $\mathbf{b}_{T}^{\mathrm{up}} \in \mathbb{R}^{d_T}$ are bias terms, $r_T \ll d_T$ is the bottleneck dimension, and $\sigma(\cdot)$ is the activation function.

The adapted tabular embedding is defined as
\begin{equation}
\label{e.adapted_tabemb}
    \widetilde{\mathbf{E}}_T
    =
    \mathbf{E}_T+\alpha_T A_T(\mathbf{E}_T),
\end{equation}
where $\alpha_T$ controls the strength of the adapter correction. 
The adapted embedding $\widetilde{\mathbf{E}}_T$ is then used as the tabular representation for multimodal fusion.

\subsection{Image Adapter}
\label{s:image_adapter}

Given an image $\mathbf{I} \in \mathbb{R}^{H \times W \times C}$, we use the ResNet-50 encoder $f_{\text{ResNet}}$~\cite{he2016deep} to extract image features. 
ResNet-50 consists of an initial convolutional stem followed by four residual layers, denoted layer1--layer4, each of which contains multiple bottleneck blocks.
The image encoder may be frozen, partially fine-tuned, or fully fine-tuned depending on the experimental setting. 
The final image embedding is written as
\begin{equation}
\label{e.imgemb}
    \mathbf{E}_I = f_{\text{ResNet}}(\mathbf{I}),
\end{equation}
where $\mathbf{E}_I \in \mathbb{R}^{d_I}$ denotes the final image representation.

We consider two adapter-based image adaptation strategies. 
The first strategy adapts the final image embedding, while the second strategy inserts convolutional adapters into selected ResNet layers at the bottleneck-block level.

\subsubsection{Embedding-level Image Adapter}
\label{s:embedding_image_adapter}

The first image adaptation strategy inserts a bottleneck adapter after the final image embedding. 
Given $\mathbf{E}_I$, the embedding-level image adapter is defined as
\begin{equation}
\label{e.img_adapter}
    A_I(\mathbf{E}_I)
    =
    W_{I}^{\mathrm{up}}
    \sigma
    \left(
    W_{I}^{\mathrm{down}} \mathbf{E}_I + \mathbf{b}_{I}^{\mathrm{down}}
    \right)
    +
    \mathbf{b}_{I}^{\mathrm{up}},
\end{equation}
where $W_{I}^{\mathrm{down}} \in \mathbb{R}^{r_I \times d_I}$, 
$W_{I}^{\mathrm{up}} \in \mathbb{R}^{d_I \times r_I}$, 
$b_{I}^{\mathrm{down}} \in \mathbb{R}^{r_I}$, 
$b_{I}^{\mathrm{up}} \in \mathbb{R}^{d_I}$, 
$r_I \ll d_I$ is the bottleneck dimension, and $\sigma(\cdot)$ is a nonlinear activation function. 
The adapted image embedding is obtained through a residual connection:
\begin{equation}
\label{e.adapted_imgemb}
    \widetilde{\mathbf{E}}_I
    =
    \mathbf{E}_I
    +
    \alpha_I A_I(\mathbf{E}_I),
\end{equation}
where $\alpha_I$ is the adapter scaling coefficient. 
This design adapts only the final pooled visual representation and does not modify intermediate ResNet feature maps.

\subsubsection{Bottleneck-level Convolutional Adapter}
\label{s:BCAdapter}

The second image adaptation strategy performs feature-map-level adaptation inside ResNet-50. 
We insert a Convolutional Adapter (ConvAdapter)~\cite{chen2024conv} after each bottleneck block within a ResNet layer, and refer to this bottleneck-level insertion strategy as Bottleneck Convolutional Adapter (BCAdapter) (see Fig.~\ref{f.framework}).

Let $\mathbf{F}_{\ell,j}$ denote the output feature map of the $j$-th bottleneck block in layer $\ell$, where $\ell \in \{1,2,3,4\}$. 
We have
\begin{equation}
    \mathbf{F}_{\ell,j} \in \mathbb{R}^{C_{\ell} \times H_{\ell} \times W_{\ell}},
\end{equation}
where $C_{\ell}$, $H_{\ell}$, and $W_{\ell}$ are the channel dimension, height, and width of the feature map in layer $\ell$, respectively.

Given $\mathbf{F}_{\ell,j}$, the ConvAdapter module in \cite{chen2024conv} is defined as
\begin{equation}
\label{e.conv_adapter}
    A_C(\mathbf{F}_{\ell,j})
    =
    \mathrm{Conv}_{1\times 1}^{\mathrm{up}}
    \left(
    \sigma
    \left(
    \mathrm{Conv}_{3\times 3}^{\mathrm{down}}(\mathbf{F}_{\ell,j})
    \right)
    \right),
\end{equation}
where $\mathrm{Conv}_{3\times 3}^{\mathrm{down}}$ maps the channel dimension from $C_{\ell}$ to $r_C$ with $r_C \ll C_{\ell}$, and $\mathrm{Conv}_{1\times 1}^{\mathrm{up}}$ maps it back from $r_C$ to $C_{\ell}$. 
The adapted feature map is obtained by residual addition:
\begin{equation}
\label{e.adapted_block_feature}
    \widetilde{\mathbf{F}}_{\ell,j}
    =
    \mathbf{F}_{\ell,j}
    +
    \alpha_{\ell,j} A_C(\mathbf{F}_{\ell,j}),
\end{equation}
where $\alpha_{\ell,j}$ is the scaling coefficient of the corresponding ConvAdapter. 
The ConvAdapter output has the same spatial resolution and channel dimension as its input, so it can be directly added to the bottleneck feature map.

\subsection{Linear Predictor}
After obtaining the adapted tabular embedding $\widetilde{\mathbf{E}}_T$ and the adapted image embedding $\widetilde{\mathbf{E}}_I$, we concatenate them to form the multimodal representation:

\begin{equation}
\label{e.fusion}
    \mathbf{Z}=\left[\widetilde{\mathbf{E}}_T,\widetilde{\mathbf{E}}_I\right]\in \mathbb{R}^{d_T+d_I}.
\end{equation}

The fused representation is then passed to a linear predictor for classification or regression:
\begin{equation}
\label{e.linear_predictor}
\mathbf{o}=W_o \mathbf{Z} + \mathbf{b}_o,
\end{equation}
where $W_o$ and $\mathbf{b}_o$ are trainable parameters.


\section{Experiments}
\label{s.exp}

\subsection{Setup}

\paragraph{Datasets}
We conduct experiments on 20 publicly available datasets from the MultaBench benchmark~\cite{arazi2026multabench}\footnote{\url{https://www.kaggle.com/chico89/datasets}}, including 12 classification tasks and 8 regression tasks. Detailed descriptions of the tabular features for each dataset are provided in \ref{a.data} and Table~\ref{T.data}. For each dataset, we randomly split the data into 80\% training and 20\% test sets. From the training set, we further reserve 20\% as a validation set for model selection and early stopping.

\paragraph{Encoder details}
We compare the proposed adapter-based methods with two common training strategies: linear probing, where only the final linear head is trained, and full fine-tuning, where all parameters of the image encoder are updated. All input images are resized to $256 \times 256$. The embedding dimension of TabPFN-v3 is set to its default value of 512, while the embedding dimension of ResNet-50 is 2048. The scaling coefficients $\alpha_T$ in \eqref{e.adapted_tabemb}, $\alpha_I$ in \eqref{e.adapted_imgemb}, and $\alpha_{\ell,j}$ in \eqref{e.adapted_block_feature} are all set to 1. The bottleneck dimensions $r_T$, $r_I$, and $r_C$ are all set to 16. 
We compare five adaptation strategies: full fine-tuning (Full), frozen ResNet with only the linear prediction head trained (Frozen), embedding-level adapter tuning (EAdapter), BCAdapter1, and BCAdapter2. 
BCAdapter1 denotes inserting ConvAdapters after each bottleneck block in layer4, while BCAdapter2 denotes inserting ConvAdapters after each bottleneck block in layer3 and layer4.

\paragraph{Implementation details}
We use accuracy (Acc.) to evaluate classification performance and mean squared error (MSE) to evaluate regression performance. All models are implemented in PyTorch and trained on a single NVIDIA 4090 GPU. Each model is trained for 100 epochs using the AdamW optimizer with an initial learning rate of $1\times 10^{-3}$ and a batch size of 64. The learning rate is decayed by a factor of 0.9 every 20 epochs. Early stopping is based on the validation loss, and training is terminated if no improvement is observed for 10 consecutive epochs. To ensure statistical robustness, all experiments are repeated with five different random seeds, and we report the mean and standard deviation of the results.

\subsection{Performance Comparison}
\label{s:performance_comparison}

Table~\ref{T.mainresults} reports the quantitative results. Overall, BCAdapter2 achieves the strongest average performance among the compared methods. On the 12 classification datasets, BCAdapter2 obtains the best mean rank of 2.08 and achieves the best result on 6 datasets, while full fine-tuning obtains a mean rank of 2.25 and achieves the best result on 5 datasets. This suggests that adapting the last two ResNet stages through convolutional adapters can achieve performance comparable to, and in some cases better than, full image encoder fine-tuning. 
For regression tasks, BCAdapter2 also achieves the best overall performance. It obtains the lowest mean rank of 1.63 and achieves the best result on 4 out of 8 datasets. In comparison, full fine-tuning obtains a mean rank of 2.25 and is best on 3 datasets. 
These results indicate that convolutional visual adapters are effective for both classification and regression tasks.

The comparison among Adapter, BCAdapter1, and BCAdapter2 further shows the importance of the adaptation location in the visual branch. The embedding-level Adapter only modifies the final pooled image representation and generally provides limited improvement. BCAdapter1 only adapts the high-level visual feature map after the last ResNet layer, while BCAdapter2 adapts both intermediate and high-level visual feature maps. The superior mean ranks of BCAdapter2 suggest that feature-map-level adaptation, especially at multiple high-level ResNet layers, is more effective than adapting only the final image embedding.

\begin{figure}[!ht]
\centering
\subfigure[Accuracy Gain]{\includegraphics[width=0.48\textwidth]{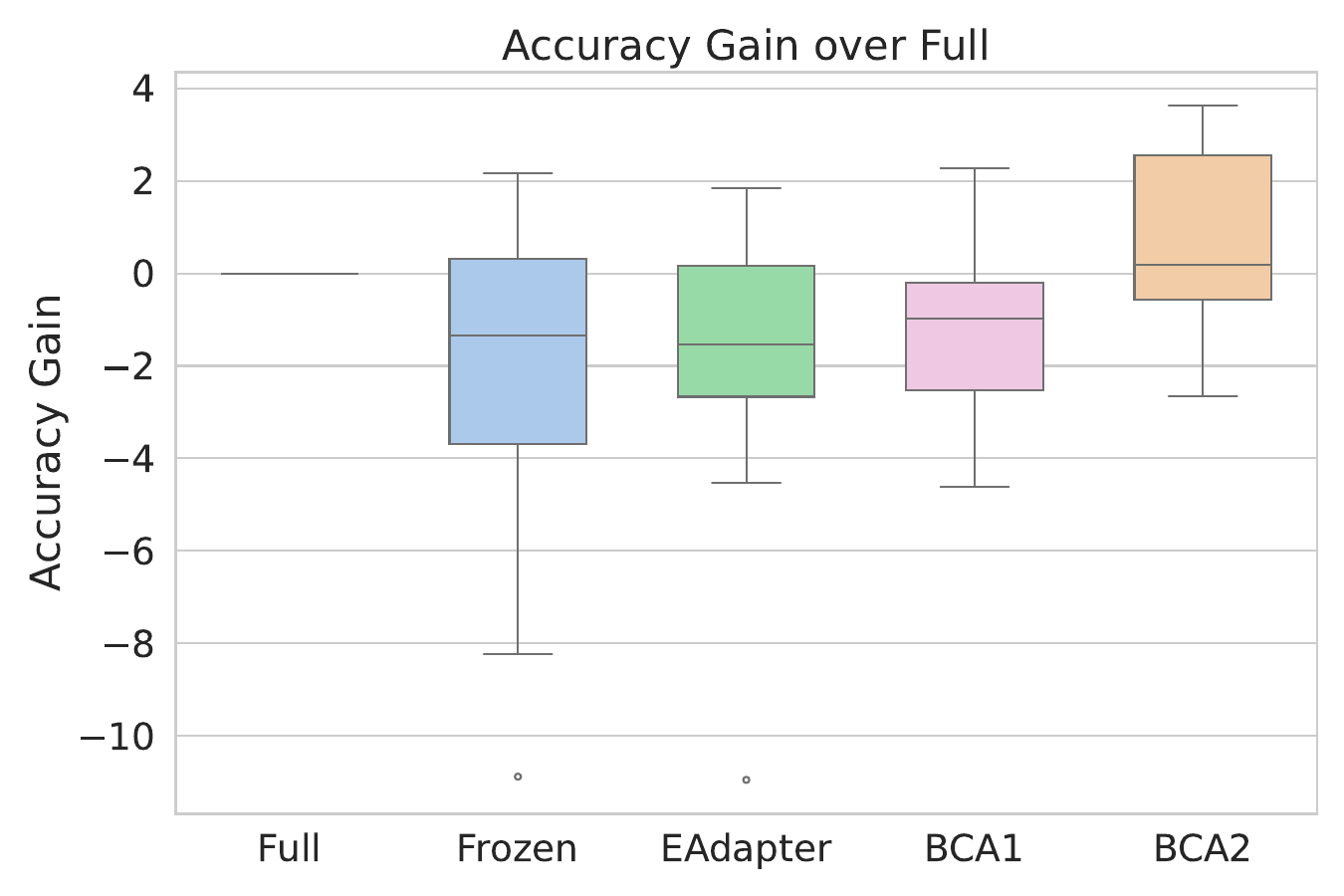}}
\subfigure[MSE Reduction]{\includegraphics[width=0.48\textwidth]{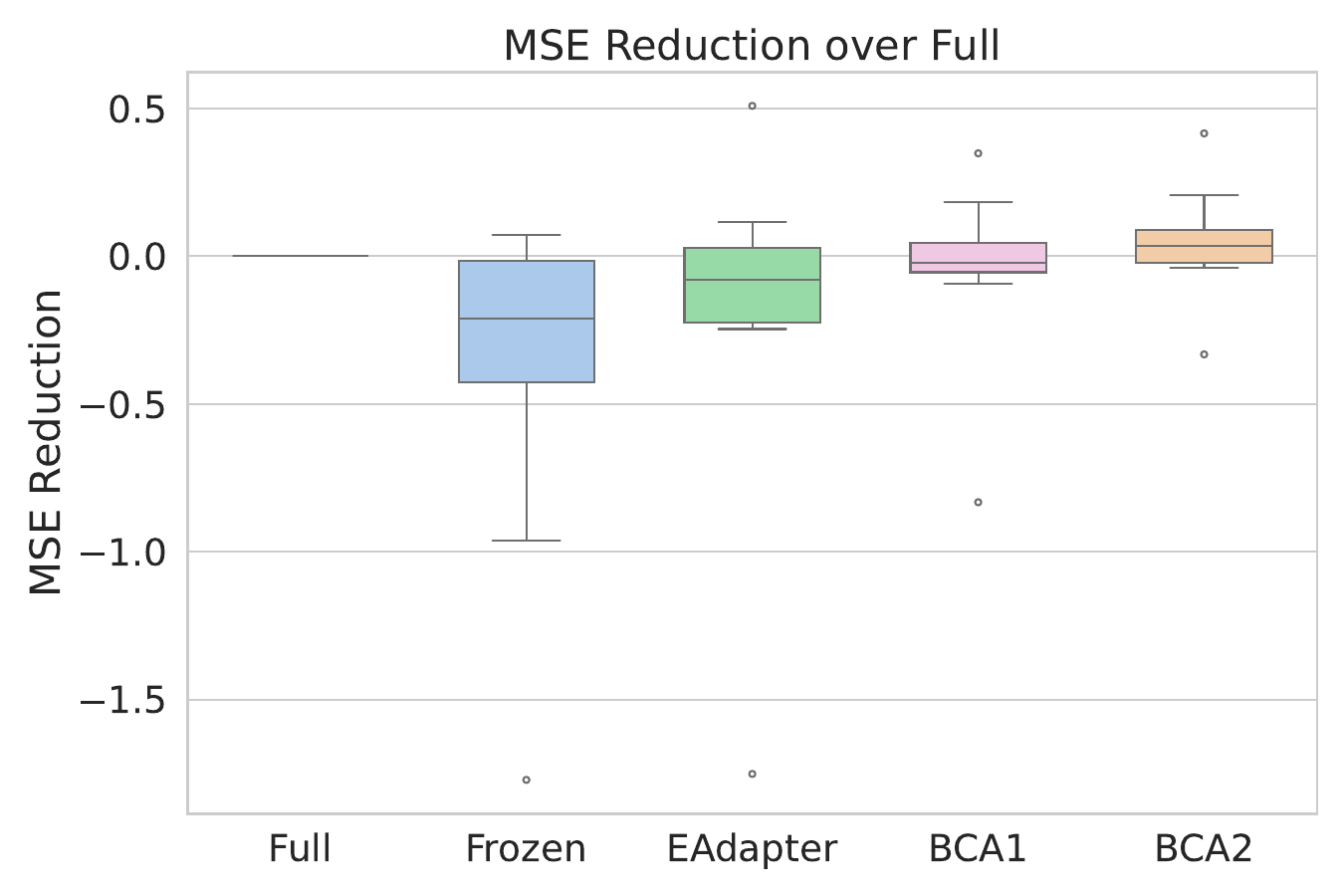}}
\caption{
Performance difference relative to full ResNet fine-tuning.
(a) Accuracy gain on 12 classification datasets.
(b) Relative MSE reduction on 8 regression datasets.
Positive values indicate better performance than full fine-tuning, while negative values indicate worse performance.
For compactness, BCA1 and BCA2 denote BCAdapter1 and BCAdapter2, respectively.
}
\label{f.gain}
\end{figure}

Fig.~\ref{f.gain} further visualizes the performance difference relative to full fine-tuning. For classification, the accuracy gain of method $m$ on dataset $d$ is computed as
\begin{equation}
    \Delta \mathrm{Acc}_{m,d}
    =
    \mathrm{Acc}_{m,d}
    -
    \mathrm{Acc}_{\mathrm{Full},d}.
\end{equation}
For regression, the relative MSE reduction is computed as
\begin{equation}
    \mathrm{MSE\ Reduction}_{m,d}
    =
    \frac{
    \mathrm{MSE}_{\mathrm{Full},d}
    -
    \mathrm{MSE}_{m,d}
    }{
    \mathrm{MSE}_{\mathrm{Full},d}
    }.
\end{equation}
For both metrics, positive values indicate that a method outperforms full fine-tuning, while negative values indicate worse performance. As shown in Fig.~\ref{f.gain}, BCAdapter2 has the most favorable distribution among the parameter-efficient methods. Its classification gains are centered closer to or above zero, and its regression reductions are generally more competitive than those of Frozen, Adapter, and BCAdapter1. These results show that BCAdapter2 can closely match or even outperform full fine-tuning on several datasets.

\begin{table}[!ht]
\renewcommand\arraystretch{1.3}
\centering
\caption{
Quantitative comparison on 20 tabular-image multimodal datasets, including 12 classification datasets and 8 regression datasets. Results are reported as mean$\pm$std. The \underline{\textbf{best result}} on each dataset is highlighted in bold and underlined. Mean Rank denotes the average rank across datasets within each task, and Best/Worst counts the number of datasets on which each method achieves the best or worst performance.
}
\label{T.mainresults}
\begin{adjustbox}{width=\textwidth}
\begin{tabular}{l|c|c|c|c|c}
\toprule[2pt]
Datasets & Full & Frozen & Adapter  & BCAdapter1 & BCAdapter2\\
\midrule[1pt]
\multicolumn{6}{c}{Classification (Acc. $\uparrow$)} \\ 
\midrule[1pt]
petfinder & 32.51$\pm$0.73 & 34.68$\pm$0.41 & 34.34$\pm$0.73 & 34.70$\pm$0.60 & \underline{\textbf{35.03$\pm$0.38}} \\

celeb-attractiveness & \underline{\textbf{81.91$\pm$0.35}} & 80.82$\pm$0.29 & 80.59$\pm$0.22 & 81.39$\pm$0.50 & 81.89$\pm$0.25 \\

glaucoma-smdg & 87.29$\pm$0.58 & 85.38$\pm$0.32 & 84.81$\pm$0.55 & 86.39$\pm$0.67 & \underline{\textbf{87.66$\pm$0.64}} \\

mammography-cmmd & \underline{\textbf{79.83$\pm$0.06}} & 79.63$\pm$0.38 & 79.79$\pm$0.08 & 78.77$\pm$1.48 & 79.48$\pm$0.28 \\

hubmap-hpa & 66.26$\pm$1.47 & 55.38$\pm$0.75 & 55.31$\pm$0.73 & 61.63$\pm$0.89 & \underline{\textbf{69.20$\pm$0.47}} \\

hateful-meme & \underline{\textbf{74.68$\pm$0.84}} & 73.88$\pm$0.27 & 73.71$\pm$0.38 & 72.20$\pm$1.51 & 73.46$\pm$1.52 \\

cbis-ddsm & 64.12$\pm$2.06 & 56.65$\pm$0.61 & 59.59$\pm$1.70 & 63.47$\pm$3.12 & \underline{\textbf{66.29$\pm$1.77}} \\

chexpert & \underline{\textbf{76.97$\pm$0.30}} & 74.55$\pm$0.16 & 74.77$\pm$0.37 & 74.72$\pm$0.90 & 76.93$\pm$0.41 \\

flower-bouquets & 35.33$\pm$2.45 & \underline{\textbf{37.33$\pm$2.81}} & 37.17$\pm$1.63 & 32.67$\pm$2.20 & 32.67$\pm$3.05 \\

zooscan-zooplankton & 88.48$\pm$0.35 & 86.89$\pm$0.15 & 86.74$\pm$0.18 & 89.25$\pm$0.09 & \underline{\textbf{92.01$\pm$0.13}} \\

csgo-skin & \underline{\textbf{39.58$\pm$3.24}} & 31.35$\pm$1.72 & 36.35$\pm$1.29 & 36.35$\pm$2.07 & 37.29$\pm$1.21 \\

justin-instagram & 85.16$\pm$0.58 & 87.00$\pm$0.12 & 85.94$\pm$0.30 & 87.44$\pm$0.28  & \underline{\textbf{87.84$\pm$0.58}} \\
\midrule[1pt]
Mean Rank & 2.25 & 3.58 & 3.83& 3.41& \textbf{2.08}\\
Best/Worst &5/2& 1/3& 0/4& 0/3& \textbf{6/1}\\
\midrule[1pt]
\multicolumn{6}{c}{Regression (MSE $\downarrow$)} \\
\midrule[1pt]
painting-price (${\times}10^{7}$)
& $1.94\pm0.0447$
& $1.81\pm0.0041$
& $1.72\pm0.0587$
& $1.58\pm0.0148$
& \underline{\textbf{1.54$\pm$0.0119}} \\

mkphoto-bots (${\times}10^{-2}$)
& \underline{\textbf{1.24$\pm$0.0419}}
& $1.50\pm0.0187$
& $1.51\pm0.0422$
& $1.39\pm0.0318$
& $1.29\pm0.0260$ \\

letterboxd-movies (${\times}10^{-1}$)
& \underline{\textbf{1.16$\pm$0.0140}}
& $1.17\pm0.0091$
& $1.19\pm0.0208$
& $1.20\pm0.0162$
& $1.18\pm0.0114$ \\

khaadi-clothes (${\times}10^{7}$)
& $1.85\pm0.1777$
& $3.63\pm0.0074$
& \underline{\textbf{0.91$\pm$0.2038}}
& $1.21\pm0.4895$
& $1.08\pm0.7141$ \\

amazon-bestseller (${\times}10^{-1}$)
& $6.11\pm0.0793$
& $7.62\pm0.0526$
& $7.61\pm0.1512$
& $6.11\pm0.1069$
& \underline{\textbf{5.98$\pm$0.2431}} \\

amazon-packages (${\times}10^{0}$)
& $9.48\pm0.0506$
& $11.55\pm0.0811$
& $10.76\pm0.1350$
& $9.88\pm0.0992$
& \underline{\textbf{9.02$\pm$0.2130}} \\

hnm-fashion (${\times}10^{1}$)
& $2.52\pm0.0942$
& $2.55\pm0.0054$
& $2.52\pm0.0155$
& $2.53\pm0.0868$
& \underline{\textbf{2.39$\pm$0.0344}} \\

mango-mass (${\times}10^{-3}$)
& \underline{\textbf{1.55$\pm$0.4751}}
& $4.29\pm0.2171$
& $4.26\pm0.2953$
& $2.84\pm0.4946$
& $2.06\pm0.2774$ \\
\midrule[1pt]
Mean Rank & 2.25  & 4.35& 3.50& 3.25& 1.63\\
Best/Worst &3/1& 0/5& 1/1& 0/1& 4/0\\
\bottomrule[2pt]
\end{tabular}
\end{adjustbox}
\end{table}

\subsection{Parameter-Efficiency Analysis}
\label{s:parameter_efficiency}

To further evaluate parameter efficiency, we analyze the relationship between predictive performance and the number of trainable parameters. Since the classification datasets have different numbers of classes, the size of the final linear prediction head varies across datasets. Therefore, instead of comparing the absolute number of trainable parameters directly, we compute the trainable parameter ratio within each dataset relative to full fine-tuning:
\begin{equation}
    \mathrm{ParamRatio}_{m,d}
    =
    \frac{
    P_{m,d}
    }{
    P_{\mathrm{Full},d}
    },
\end{equation}
where $P_{m,d}$ is the number of trainable parameters of method $m$ on dataset $d$, and $P_{\mathrm{Full},d}$ is the corresponding number for full fine-tuning. The final parameter ratio of each method is obtained by averaging $\mathrm{ParamRatio}_{m,d}$ over all datasets.

\begin{figure}[!ht]
    \centering
    \includegraphics[width=0.95\linewidth]{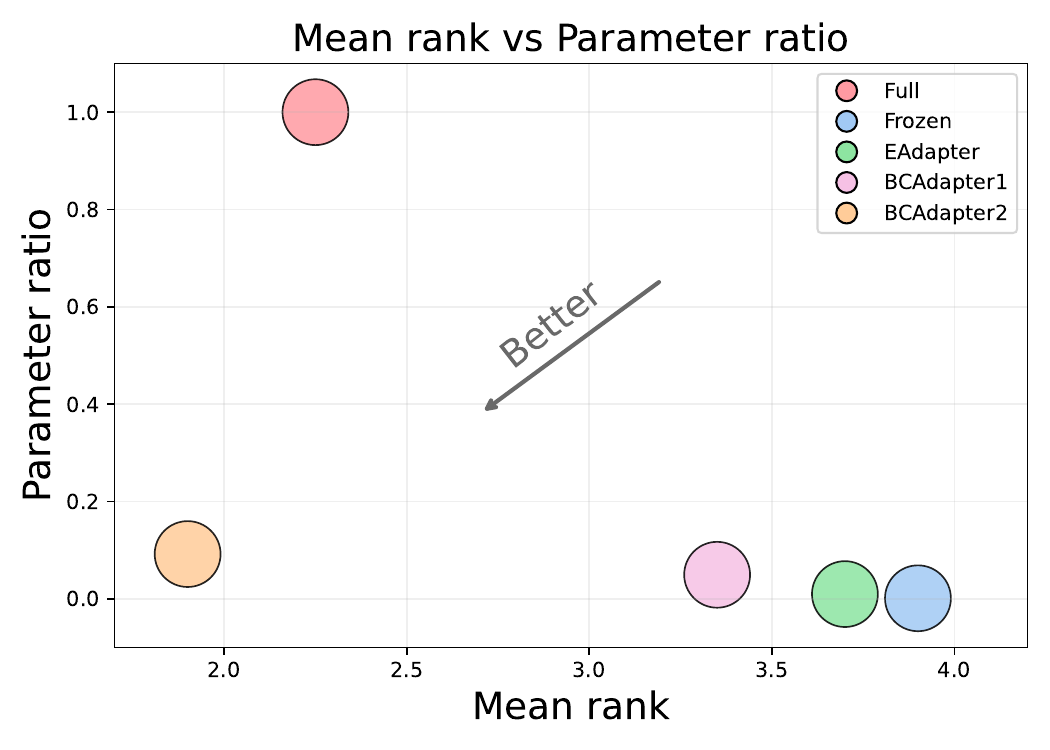}
    \caption{
    Performance--parameter efficiency trade-off across 20 tabular-image multimodal datasets. 
    The x-axis denotes the unified mean rank, where a lower rank indicates better predictive performance. 
    The y-axis denotes the average trainable parameter ratio relative to full fine-tuning. 
    Methods closer to the bottom-left corner achieve a better trade-off between predictive performance and parameter efficiency.
    }
    \label{f.bubble}
\end{figure}

To compare classification and regression tasks under a unified criterion, we use mean rank. 
Fig.~\ref{f.bubble} visualizes the trade-off between unified mean rank and trainable parameter ratio. Full fine-tuning achieves a strong mean rank but requires the largest number of trainable parameters. In contrast, Frozen and Adapter use substantially fewer trainable parameters, but their mean ranks are worse, indicating limited adaptation capacity. BCAdapter2 achieves the most favorable trade-off. It obtains a competitive mean rank while maintaining a much lower trainable parameter ratio than full fine-tuning. This confirms that adapting high-level ResNet feature maps with convolutional adapters can preserve strong predictive performance while substantially reducing the number of trainable parameters.

\subsection{Ablation Study}
\label{s:ablation}

\begin{figure}[!ht]
\centering
\subfigure[Accuracy gain]{
    \includegraphics[width=0.45\textwidth]{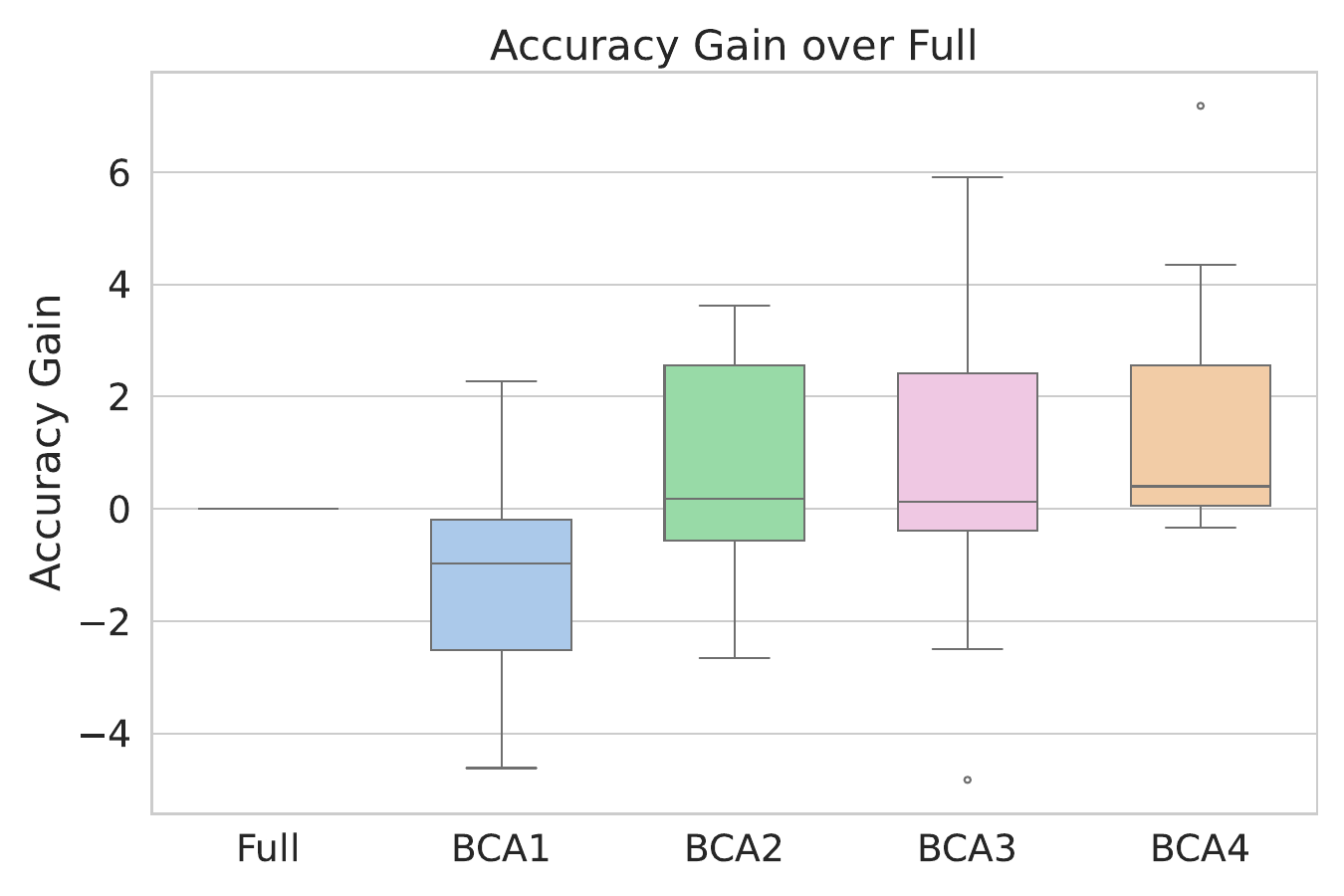}}
\subfigure[MSE reduction]{
    \includegraphics[width=0.45\textwidth]{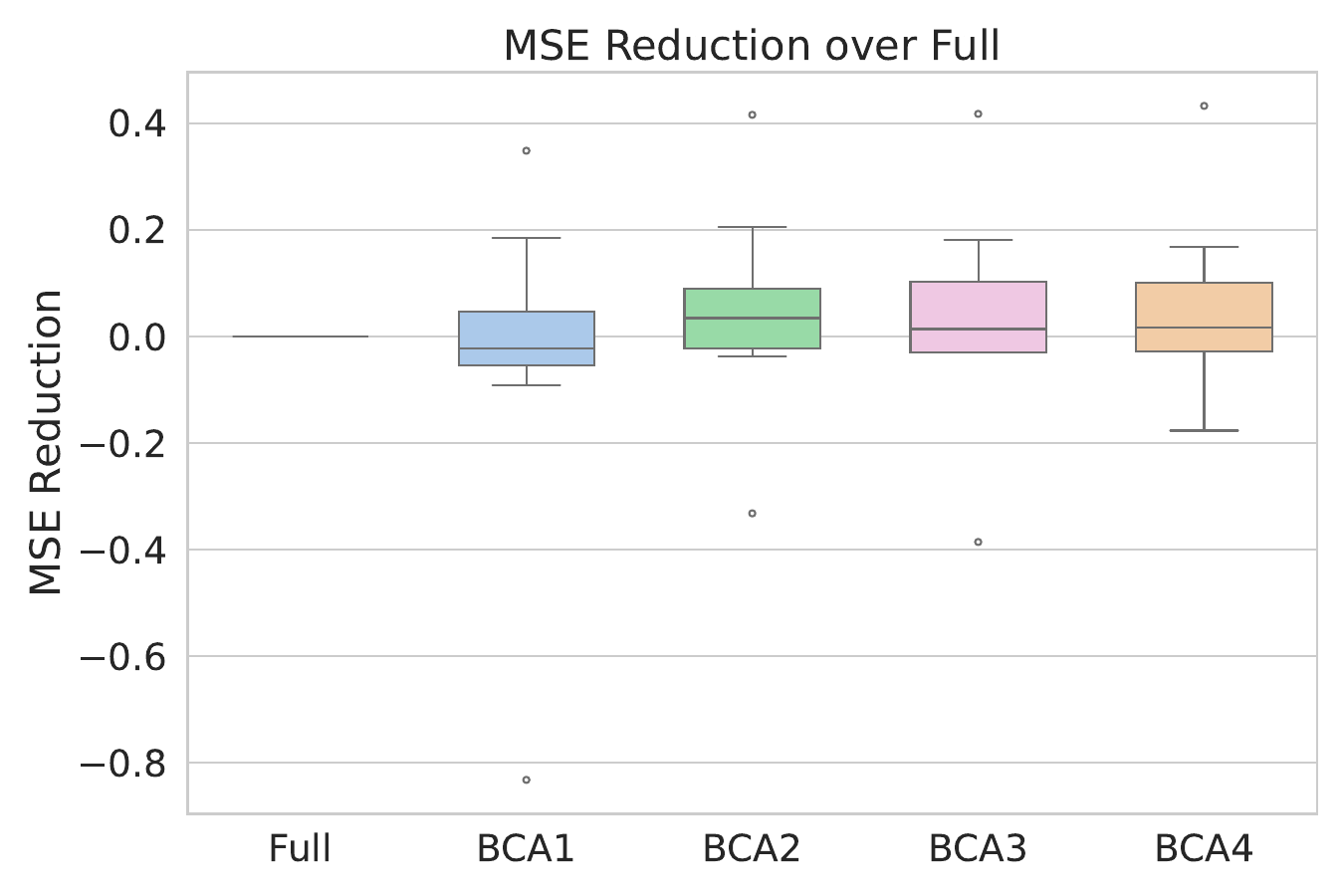}}
\subfigure[Average memory ratio]{
    \includegraphics[width=0.45\textwidth]{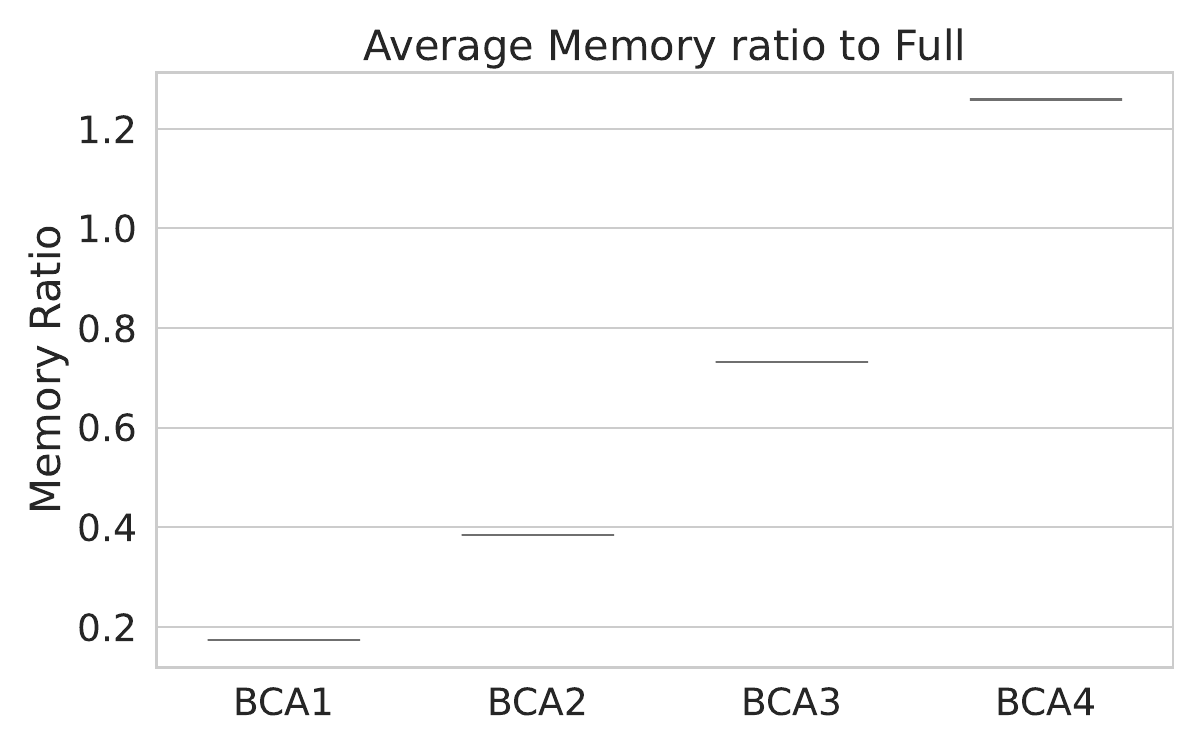}}
\subfigure[Average parameter ratio]{
    \includegraphics[width=0.45\textwidth]{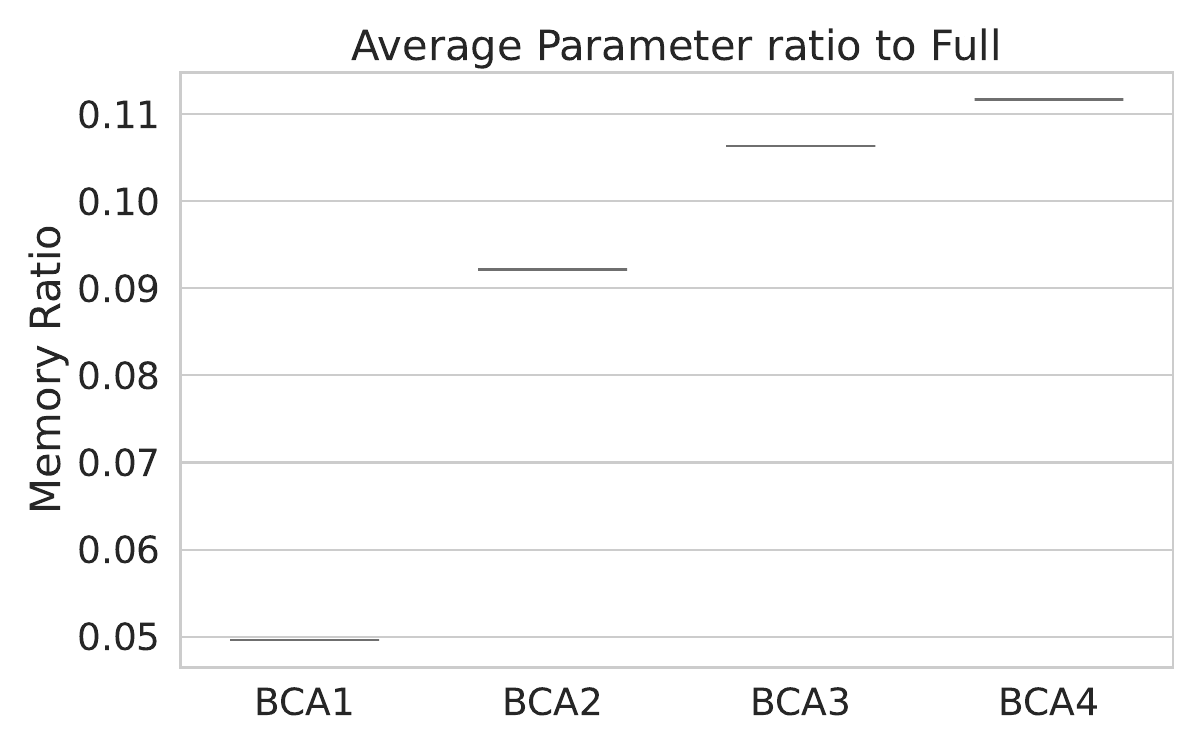}}
\caption{
Ablation study on the number of ResNet layers equipped with BCAdapters. 
For compactness, BCA1, BCA2, BCA3, and BCA4 denote BCAdapter1, BCAdapter2, BCAdapter3, and BCAdapter4,respectively.
(a) and (b) show performance improvement over full fine-tuning, measured by accuracy gain and MSE reduction. 
(c) and (d) show the GPU memory ratio and trainable parameter ratio relative to full fine-tuning.
}
\label{f.ablation}
\end{figure}

We further conduct an ablation study to analyze how the number of ResNet layers equipped with BCAdapters affects performance and efficiency. 
BCAdapter3 and BCAdapter4 denote inserting ConvAdapters into the last three and all four ResNet layers, respectively. 
All variants are compared with full fine-tuning.

Fig.~\ref{f.ablation} summarizes the ablation results and more quantitative results can be found in Table~\ref{T.ablationresults}, Table~\ref{T.param_ratio}, and Table~\ref{T.memory_ratio} in \ref{a.ablation}. 
For classification, BCAdapter1 generally underperforms full fine-tuning, whereas BCAdapter2, BCAdapter3, and BCAdapter4 achieve more competitive accuracy gains. 
This suggests that adapting only the last ResNet layer is not sufficient, while adapting the last two or more layers provides stronger visual adaptation. 
For regression, BCAdapter2--BCAdapter4 also tend to obtain positive MSE reductions over full fine-tuning, although the differences among these variants are relatively small.

The cost analysis reveals a clear trade-off. 
As more ResNet layers are equipped with BCAdapters, the trainable parameter ratio gradually increases but remains far below that of full fine-tuning. 
However, the GPU memory cost increases much more rapidly. 
In particular, BCAdapter4 can require more memory than full fine-tuning, even though it uses fewer trainable parameters. 
This shows that reducing trainable parameters does not necessarily reduce practical training memory.

This phenomenon is mainly caused by the memory consumption of ResNet training.
Trainable parameter count only measures the number of learnable weights, whereas training memory is also affected by the storage of intermediate feature maps and their gradients~\cite{chen2016training}.
When BCAdapters are inserted into earlier ResNet layers, they operate on feature maps with larger spatial resolutions, which increases activation storage and backward computation costs.
Moreover, because trainable adapters are interleaved with frozen bottleneck blocks, gradients still need to propagate through later frozen blocks to update earlier adapters.
Therefore, parameter-efficient adaptation may still introduce substantial memory overhead even though it trains only a small number of parameters~\cite{mercea2024time}.

Overall, BCAdapter4 achieves the strongest raw performance, but its improvement over BCAdapter2 is limited while its memory cost is substantially higher. 
We therefore select BCAdapter2 as the default design. 
It adapts the last two ResNet layers, achieves performance competitive with or better than full fine-tuning, and avoids the excessive memory overhead caused by adapting early layers.

\section{Conclusion}
\label{s.conclusion}
This paper studies parameter-efficient adaptation for tabular-image multimodal learning. 
We propose TI-Adapter, a modality-specific fine-tuning framework that adapts the tabular and image branches with lightweight adapters. 
In our implementation, the pretrained TabPFN encoder is kept frozen and equipped with a tabular embedding adapter, while the ResNet image encoder is adapted through embedding-level and bottleneck-level adapters instead of full fine-tuning.

Experiments on 20 tabular-image datasets show that TI-Adapter achieves performance competitive with full fine-tuning while using substantially fewer trainable parameters. 
The ablation study further shows that inserting adapters into more ResNet layers can improve raw performance, but may also increase GPU memory cost. 
Overall, these results indicate that modality-specific adapter tuning provides a practical alternative to full fine-tuning for tabular-image multimodal learning.



\appendix
\section{Dataset Description}
\label{a.data}

We use 20 datasets in our experiments, including 12 classification datasets and 8 regression datasets. 
Table~\ref{T.data} summarizes the main statistics of these datasets.

\begin{table}[!ht]
\renewcommand\arraystretch{1.2}
\footnotesize
\centering
\caption{
Dataset statistics. 
\#Samples denotes the total number of samples in each dataset, \#Features denotes the number of tabular features, and \#Classes denotes the number of target classes for classification datasets.
}
\label{T.data}
\begin{tabular}{c|c|c|c}
\toprule[2pt]
Dataset  & \#Sample  & \#Feature & \#Class  \\
\midrule[1pt]
\multicolumn{4}{c}{Classification} \\
\midrule[1pt]
petfinder  & 14652 &  22 & 8   \\
 \hline
celeb-attractiveness & 99999 & 40 & 2 \\
 \hline
glaucoma-smdg  & 12449 & 9 & 3  \\
 \hline
mammography-cmmd  & 5202 & 5 & 2  \\
\hline
hubmap-hpa  & 12581 &  7 & 10   \\
 \hline
hateful-meme & 9000 & 21 & 2 \\
 \hline
cbis-ddsm  & 1696 & 9 & 4  \\
 \hline
chexpert  & 46437 & 18 & 3  \\
\hline
flower-bouquets  & 600 &  5 & 5   \\
 \hline
zooscan-zooplankton & 100000 & 29 & 10 \\
 \hline
csgo-skin  & 956 & 5 & 10  \\
 \hline
justin-instagram  & 10319 & 7 & 5  \\
\midrule[1pt]
\multicolumn{4}{c}{Regression} \\
\midrule[1pt]
painting-price & 12369 & 248 & 1 \\
\hline
mkphoto-bots & 13748 & 9 & 1 \\
\hline
letterboxd-movies & 12564 & 27 & 1 \\
\hline
khaadi-clothes & 400 & 4 & 1 \\
\hline
amazon-bestseller & 3488 & 5 & 1 \\
\hline
amazon-packages & 46398 & 3 & 1 \\
\hline
hnm-fashion & 104072 & 14 & 1 \\
\hline
mango-mass & 546 & 3 & 1 \\
 \bottomrule[2pt]
\end{tabular}
\end{table}

\section{Additional Ablation Results}
\label{a.ablation}

\subsection{Performance of Adapter-Depth Variants}
\label{a.adapter_depth}

Table~\ref{T.ablationresults} reports the complete results of BCAdapter3 and BCAdapter4 on all 20 datasets. 
These two variants insert ConvAdapters into the last three ResNet layers and all four ResNet layers, respectively. 
The results provide additional evidence for the adapter-depth ablation in Section~\ref{s:ablation}. 
Overall, increasing the number of adapted layers can further improve performance on some datasets, especially for classification tasks such as \textit{flower-bouquets}, \textit{csgo-skin}, and \textit{cbis-ddsm}. 
However, the gains are not consistent across all datasets and are relatively limited for several regression tasks. 
Together with the cost analysis in Tables~\ref{T.param_ratio} and~\ref{T.memory_ratio}, these results support our choice of BCAdapter2 as the default configuration.

\begin{table}[!ht]
\footnotesize
\renewcommand\arraystretch{1.3}
\centering
\caption{
Performance of deeper adapter-depth variants on 20 datasets. 
BCAdapter3 inserts ConvAdapters into the last three ResNet layers, while BCAdapter4 inserts ConvAdapters into all four ResNet layers. 
Classification results are reported by accuracy, and regression results are reported by MSE. 
}
\label{T.ablationresults}
\begin{tabular}{lcc}
\toprule[2pt]
Dataset & BCAdapter3 & BCAdapter4 \\
\midrule[1pt]
\multicolumn{3}{c}{Classification (Acc. $\uparrow$)} \\ 
\midrule[1pt]
petfinder & $35.12\pm1.20$ & $34.67\pm0.52$ \\
celeb-attractiveness & $82.04\pm0.35$ & $82.19\pm0.13$ \\
glaucoma-smdg & $87.42\pm0.53$ & $87.37\pm0.27$ \\
mammography-cmmd & $79.67\pm0.33$ & $79.60\pm0.62$ \\
hubmap-hpa & $72.17\pm1.04$ & $73.44\pm0.55$ \\
hateful-meme & $73.62\pm0.46$ & $74.67\pm0.75$ \\
cbis-ddsm & $65.35\pm2.30$ & $67.88\pm1.83$ \\
chexpert & $76.84\pm0.20$ & $77.12\pm0.42$ \\
flower-bouquets & $30.50\pm1.80$ & $35.00\pm2.04$ \\
zooscan-zooplankton & $92.74\pm0.20$ & $92.83\pm0.21$ \\
csgo-skin & $37.08\pm4.58$ & $40.10\pm2.19$ \\
justin-instagram & $87.51\pm0.30$ & $86.68\pm0.30$ \\
\midrule[1pt]
\multicolumn{3}{c}{Regression (MSE $\downarrow$)} \\
\midrule[1pt]
painting-price (${\times}10^{7}$)
& $1.94\pm0.0175$
& $1.81\pm0.0119$ \\
mkphoto-bots (${\times}10^{-2}$)
& $1.59\pm0.0222$
& $1.62\pm0.0143$ \\
letterboxd-movies (${\times}10^{-1}$)
& $1.28\pm0.0108$
& $1.27\pm0.0233$ \\
khaadi-clothes (${\times}10^{7}$)
& $1.20\pm0.1248$
& $1.19\pm0.6322$ \\
amazon-bestseller (${\times}10^{-1}$)
& $6.23\pm0.0201$
& $6.35\pm0.0148$ \\
amazon-packages (${\times}10^{0}$)
& $8.75\pm0.0756$
& $8.98\pm0.0169$ \\
hnm-fashion (${\times}10^{1}$)
& $2.40\pm0.0393$
& $2.32\pm0.0258$ \\
mango-mass (${\times}10^{-3}$)
& $2.15\pm0.4200$
& $1.82\pm0.2101$ \\
\bottomrule[2pt]
\end{tabular}
\end{table}

\subsection{Parameter and Memory Cost}
\label{a.cost}

Tables~\ref{T.param_ratio} and~\ref{T.memory_ratio} report the trainable parameter counts and peak GPU memory usage of different adaptation strategies. 
For each dataset, the parameter and memory ratios are computed relative to the corresponding full fine-tuning setting and then averaged across datasets. 
All adapter-based methods substantially reduce the number of trainable parameters: EAdapter uses less than 1\% of the trainable parameters of full fine-tuning, while BCAdapter2 uses only 9.22\%. 
However, the memory ratio does not decrease proportionally with the parameter ratio. 
For example, BCAdapter4 uses only 11.17\% of the trainable parameters but reaches 126.0\% of the peak GPU memory of full fine-tuning. 
This further shows that parameter efficiency and memory efficiency are not always aligned.

\begin{table}[!ht]
\footnotesize
\renewcommand\arraystretch{1.2}
\centering
\caption{
Trainable parameter numbers and ratios relative to full fine-tuning. 
The parameter ratio is computed for each dataset relative to the corresponding full fine-tuning setting and then averaged across datasets.
}
\label{T.param_ratio}
\begin{tabular}{lccc}
\toprule
Method & Min Params & Max Params & Mean Param Ratio (\%) \\
\midrule
Full      & 23.51M & 23.57M & 100.0 \\
Frozen     & 6.15K  & 61.45K & 0.13 \\
EAdapter   & 208.93K & 264.23K & 0.99 \\
BCAdapter1 & 1.14M & 1.20M & 4.96 \\
BCAdapter2 & 2.14M & 2.20M & 9.22 \\
BCAdapter3 & 2.48M & 2.53M & 10.63 \\
BCAdapter4 & 2.60M & 2.66M & 11.17 \\
\bottomrule
\end{tabular}
\end{table}

\begin{table}[!ht]
\footnotesize
\renewcommand\arraystretch{1.2}
\centering
\caption{Peak GPU memory usage and ratios relative to full fine-tuning. The memory ratio is computed within each dataset and then averaged across datasets.}
\label{T.memory_ratio}
\begin{tabular}{lccc}
\toprule
Method & Min Peak Mem (MB) & Max Peak Mem (MB) & Mean Mem Ratio (\%) \\
\midrule
Full      & 5808 & 5834 & 100.0 \\
Frozen     & 984  & 1007 & 17.0 \\
EAdapter   & 987  & 1010 & 17.1 \\
BCAdapter1 & 997  & 1022 & 17.3 \\
BCAdapter2 & 2225 & 2252 & 38.4 \\
BCAdapter3 & 4248 & 4276 & 73.2 \\
BCAdapter4 & 7316 & 7342 & 126.0 \\
\bottomrule
\end{tabular}
\end{table}

\bibliographystyle{elsarticle-num} 
\bibliography{references}

\end{document}